%% file: main.tex
\documentclass[runningheads]{llncs}
\usepackage[T1]{fontenc}
\input{math_commands.tex}

\usepackage{graphicx}
\usepackage{hyperref}
\usepackage{url}
\usepackage{algorithm}
\usepackage{algpseudocode}
\usepackage{multirow}

\usepackage{color}

\usepackage{lipsum}

\begin{document}
\title{GDC-- Generalized Distribution Calibration for Few-Shot Learning}
\titlerunning{GDC-- Generalized Distribution Calibration for Few-Shot Learning}
%

\author{Shakti Kumar, Hussain Zaidi}
\authorrunning{S. Kumar, H. Zaidi}

%
\institute{Vanguard Center for Analytics and Insights\\
\email{\{shakti\_kumar, hussain\_zaidi\}@vanguard.com}}

%
\maketitle              
\begin{abstract}
Few shot learning is an important problem in machine learning as large labelled datasets take considerable time and effort to assemble. Most few-shot learning algorithms suffer from one of two limitations--- they either require the design of sophisticated models and loss functions, thus hampering interpretability; or employ statistical techniques but make assumptions that may not hold across different datasets or features. Developing on recent work in extrapolating distributions of small sample classes from the most similar larger classes, we propose a Generalized sampling method that learns to estimate few-shot distributions for classification as weighted random variables of all large classes. We use a form of covariance shrinkage to provide robustness against singular covariances due to overparameterized features or small datasets. 
We show that our sampled points are close to few-shot classes even in cases when there are no similar large classes in the training set.
Our method works with arbitrary off-the-shelf feature extractors and outperforms existing state-of-the-art on miniImagenet, CUB and Stanford Dogs datasets by 3\% to 5\% on 5way-1shot and 5way-5shot tasks and by 1\% in challenging cross domain tasks.

\end{abstract}

\section{Introduction}
Few-shot learning (FSL) refers the problem of learning from datasets where only a limited number of examples (typically, one to tens per class or a problem in general) are available for training a machine learning model. FSL has gained importance over the years since obtaining large labelled datasets requires a significant investment of time and resources. There is a rich history of research into various methodologies for FSL, two primary ones being model development approaches and representation learning. \cite{DBLP:journals/csur/WangYKN20}. 

Model development approaches aim at capturing the data distribution so that new points can be sampled to improve few-shot classification accuracy \cite{pmlr-v119-park20b,DBLP:journals/corr/abs-1909-12220,Chen_2019,Zhang_2019_ICCV}. They have typically relied on complex models and loss functions to understand data from only a few examples, which limits the interpretability of the model and hampers generalization.
Representation learning, on the other hand, aims at identifying feature transformations which can allow simple statistical techniques like nearest neighbor and bayesian classification generalize on few-shot tasks for novel classes \cite{DBLP:journals/corr/abs-2003-04390,Xue_Wang_2020}. Starting from early works \cite{DBLP:conf/cvpr/MillerMV00} and building to recent methodologies \cite{DBLP:conf/iclr/YangLX21,9455950}, representation learning strategies have relied on simple statistical assumptions that may not hold across diverse datasets.

Recently, \cite{DBLP:conf/iclr/YangLX21} showed that classes which are semantically similar in meaning are also correlated in the means and covariances of their feature distributions. 
They used the statistics of classes with plentiful datapoints (called base classes) to learn the distribution of classes with a few datapoints (called novel classes). Despite some limiting assumptions, their method, called Distribution Calibration, outperformed much more complex models that relied on non-parametric optimization and generative models \cite{pmlr-v119-park20b,DBLP:journals/corr/abs-1909-12220,Chen_2019,Zhang_2019_ICCV}. Building on this idea and inspired by traditional statistics, we present a rigorous generalized sampling method for Few-Shot Learning, called GDC that outperforms existing state-of-the-art classification accuracy without introducing additional statistical assumptions or complex generative models.
Our main contributions are: 
\begin{enumerate}
    \item Introducing a principled approach to estimate novel class mean and covariance from the moments of a random variable weighted by the distance between the novel and the base classes, 
    \item Incorporating the statistical technique of variance shrinkage, which not only helps increase the accuracy but also stabilizes covariance estimation in cases when the feature extractor is over-parametrized (a common occurrence in modern deep learning models),
    \item Extending the applicability of the statistical sampling approach to arbitrary feature extractors by introducing general Gaussianization transformations, 
    \item Presenting a single scaling parameter in Euclidean distance weighting that mitigates the need to search among Euclidean, Mahalanobis, and generalized distances for novel class estimation. 
\end{enumerate}
Combined, our contributions put statistical sampling approach on a sound foundation, close open questions in earlier research, and demonstrate $3\%$ to $5\%$ improvement over competitive states-of-the-art for 5way-1shot and 5way-5shot tasks for miniImageNet \cite{DBLP:conf/iclr/RaviL17}, CUB \cite{WelinderEtal2010}, StanfordDogs \cite{KhoslaYaoJayadevaprakashFeiFei_FGVC2011}, highest level classes of tieredImagenet \cite{ren2018metalearning} and 1\% improvement Cross Domain 5way-1shot task of miniImagenet $\xrightarrow[]{}$ CUB.







\section{Related Works}
Learning good features or manipulating the features to help generalize few-shot tasks is an active research area \cite{9010298,hou2019cross,Li2019FindingTF}. \cite{DBLP:conf/cvpr/MillerMV00} proposed a congealing process which learned a joint distribution among all available classes. This distribution could then be used as a prior knowledge for constructing efficient few-shot classifiers. \cite{DBLP:journals/corr/abs-2003-04390} showed that by pre-training first on entire base classes, few shot classification accuracies on novel classes could be improved with a simple meta-learning on nearest-centroid based classification algorithm. Their main focus was on the pretraining methods which could improve a cosine based classifier on the centroids of the extracted features. GDC can be applied on top of these feature extractor techniques to further explore improvements. 

To correct bias in centroid or prototype estimations, several rectification methods were proposed--- RestoreNet \cite{Xue_Wang_2020} transforms the feature space to move the images closer to their true centroids. Our proposed method approximates this transformation with scaled Euclidean distances and weighted random variables for novel classes, without any additional learnable parameters. \cite{zhang2021prototype} proposed a 4-step method consisting of learning base class details as priors and then using these priors for correcting bias in novel prototype estimation. They then jointly fine tuned the feature extractor and bias corrector. Our method does not have this multi-step process and works with off-the-shelf feature extractors. \cite{liu2020prototype} attempted to reduce the bias in distance estimation through reducing intra-class bias by label propagation and cross-class bias through shifting features. Their method works in the transductive setting where entire data, including the query set is consumed without label information. Our method does not need additional unlabelled data from the query set. Again their method improves upon Prototypical Networks \cite{DBLP:conf/nips/SnellSZ17} whereas we show that our method can be applied with any feature extractor.

GDC can be broadly categorized as a data augmentation method. Several methods have been proposed earlier in this space. \cite{DBLP:journals/corr/abs-1902-09884} learn few-shot tasks by randomly labelling a subset of images and then augmenting this subset with different techniques like random crops, flips etc. \cite{pmlr-v119-park20b} tried to transfer variance between different classes in order to simulate the query examples that can be encountered during test. Other works like \cite{DBLP:journals/corr/abs-1909-12220,DBLP:conf/cvpr/LiuSHDL20} utilized the intra-class variance to perform augmentation. These methods leverage complex neural networks with large number of learnable parameters to generate new examples. Our method does not require any additional learnable parameter and uses simple statistical techniques while still outperforming all previous deep learning methods.

Closest to GDC, \cite{DBLP:conf/iclr/YangLX21} estimated the novel class distributions based on their similarity with the base classes. Their method implicitly assumed that the base classes were semantically independent of each other when constructing covariance estimates, did not consider the similarity strength between the base and novel classes when estimating novel class statistics, and could not be applied to arbitrary off-the-shelf feature extractors (with activation functions different from \texttt{relu}) and large feature dimensions often capable of producing ill-defined covariances. Our method does not make independence assumptions, leverages similarity information in the base classes, and can be applied to any off-the-shelf feature extractor.

\section{Proposed Approach}

\subsection{Problem Definition}
\label{ProblemDefinition}
Few-shot classification problems are defined as $N$way-$K$shot classification tasks $\gT$ \cite{DBLP:journals/corr/VinyalsBLKW16} where given a small support set $\sS$ of features $\rvxtilda$ and labels $y$, $\sS=\{(\rvxtilda_i, y_i)\}_{i=1}^{N\times K}, \rvxtilda_i \in \sR^d, y_i \in \sC$, consisting of $K$ points from $N$ classes, the model should correctly classify a query set $\sQ=\{(\rvxtilda_i, y_i)\}_{i=N\times K+1}^{N\times K + N \times q}$ with $q$ points from each of the $N$ classes in the support set.
The entire dataset $\sD$, is divided into $\sC_b$ base, $\sC_v$ validation and $\sC_n$ novel classes such that $\sC_b\cap \sC_v\cap \sC_n = \phi$ and $\sC_b\cup \sC_v\cup \sC_n = \sC$. The goal is to train a model with tasks $\gT$ sampled from $\sC_b$ and use $\sC_v$ for hyperparameter tuning, where each task $\gT$ is an $N$way-$K$shot classification problem on $N$ unique classes of the set under consideration, for example base, validation set $\sC_b, \sC_v$ here. The performance of few-shot learning algorithms is reported as the average accuracy on the query set $\sQ$ of tasks $\gT$ sampled from $\sC_n$.

\subsection{Algorithm}
\label{sec:algorithm}
Our proposed methodology, GDC, is outlined in Algorithm \ref{alg:cap}. In the following subsections, we incrementally go through the steps of GDC in detail. 

\subsubsection{Gaussianization of the Data}
\label{GaussianizationData}
Following \cite{DBLP:conf/iclr/YangLX21}, our sampling methodology assumes that the input features follow a multivariate normal distribution. There are many methods of data gaussianization like Tukey's Ladder of Powers \cite{DBLP:books/lib/Tukey77}, Yeo-Johnson Transformation \cite{Weisberg01yeo-johnsonpower}, and Iterative Gaussianization \cite{DBLP:conf/nips/ChenG00}. In our experiments, we observed that Tukey's Ladder of Powers outperformed other methods, but the fractional powers and log transform could only be applied to non-negative features (feature extractors which have \texttt{relu} and equivalent activation functions in their final layers). Expanding the applicability of the sampling method to arbitrary feature extractors and activation functions in deep learning models, we make a choice on the transformation of input features denoted by random variable $\rvx$ as per \eqref{TukeyorYJ} below,
\begin{equation}
\hat{\rvx} = \left\{
        \begin{array}{ll}
            tukey(\rvx) & if \; \rvx \geq 0\;\;always\\
            yeo\_johnson(\rvx) & otherwise
        \end{array}
    \right.
\label{TukeyorYJ}
\end{equation}
where $tukey(\rvx)$ and $yeo\_johnson(\rvx)$ are defined as,
\begin{equation}
tukey(\rvx) = \left\{
        \begin{array}{ll}
            \rvx^\beta &\beta \neq 0 \\
            \log(\rvx) &\beta = 0 \\
        \end{array}
    \right.
\label{Tukey}
\end{equation}

\begin{equation}
yeo\_johnson(\rvx) = \left\{
        \begin{array}{ll}
            ((\rvx + 1)^\beta-1)/\beta &\beta \neq 0, \rvx \geq 0 \\
            \log(\rvx + 1) &\beta = 0, \rvx \geq 0 \\
            -[(-\rvx + 1)^{2-\beta}-1]/(2-\beta) &\beta \neq 2, \rvx < 0 \\
            -\log(-\rvx + 1) &\beta = 2, \rvx < 0 \\
        \end{array}
    \right.
\label{YJ}
\end{equation}

\subsubsection{Proposed Random Variable}
\label{Proposed_Random_Variable}
We extrapolate the distribution of a given novel class as a weighted average of the distributions of $k$ closest base classes. Formally, if $\rvxtilda \in \sR^d$ is a $d$ dimensional support point from a novel class, and $\rmX_i \in \sR^d$ is a random variable denoting points of base class $i$, then we compose a random variable $\rmX'$ representing our estimate of that novel class as
\begin{equation}
    \rmX' = \frac{\rvxtilda + \sum_{i \in \sS_k}w_i\rmX_i}{1+\sum_{i \in \sS_k}{w_i}},
\label{weighted-rv}
\end{equation}
where $w_i$ are the weights assigned to the $k$ closest base classes in $\sS_k$.
The associated mean and the covariance of this random variable are (since $\rvxtilda$ is a constant vector, it does not affect the covariance in $\mSigma'$),
\begin{equation}
\rvmu' = \frac{\rvxtilda + \sum_{i \in \sS_k}w_i\rvmu_i}{1+\sum_{i \in \sS_k}{w_i}} \;\;\;\;\;\; \mSigma' = cov(\rmX')
\label{mean_prime_sigma_prime}
\end{equation}
where $\rvmu_i=\E[\rmX_i]$.
There are many ways of estimating $w_i$. One of the simplest is to look at the distance of the novel point from the base classes. In particular, we find the $k$ closest base classes to $\rvxtilda$ in $\sS_k$,
\begin{flalign}
        d_i = ||\rvxtilda - \rvmu_i||^2, i \in \sC_b \label{find_di}\\
        \sS_k = \{i |-d_i \in topk(-d_i), i \in \sC_b \} \label{find_sk}
\end{flalign}
Based on $d_i$\footnote{for alternative distance formulations, see Supplementary section \ref{Effect_of_Distances}}, we construct the weights $w_i$ of each base class in $\sS_k$ as,
\begin{equation}
    w_i = \frac{1}{1+d_i^m} , i \in \sS_k
\label{weights}
\end{equation}

where $m$ is a hyperparameter that helps in decaying $w_i$ as a function of $d_i$ and gives us control in the relative weights of the classes in $\sS_k$. This form of weighted variable estimation is reminiscent of (though not the same as) inverse distance weighting, which is widely used to estimate unknown functions at interpolated points \cite{10.1145/800186.810616}, \cite{2004CompM..33..299L}.
It is worth noting that $\lim_{d_i\to0} w_i = 1$. Hence as the base class $i$ moves closer to the support point $\rvxtilda$, it gains increasing weight until $\rvmu_i$ overlaps with $\rvxtilda$, at which point the weight is $1$, same as DC method. 

\subsubsection{Shrinking the Covariance}
\label{Shrinking_the_Covariance}
When the number of data points in base class $\rmX_i$ is less than the feature dimensions, i.e. $|\rmX_i| < d$, $cov(\rmX_i) = \mSigma_i$ is in general non-invertible, which prohibits constructing a normal distribution with $\mSigma'$ and poses a serious limitation since most off-the-shelf deep learning feature extractors have large feature dimensions \cite{DBLP:journals/corr/ZagoruykoK16}, \cite{szegedy2015rethinking}. We propose a variant of covariance shrinkage 
\cite{1980PatRe..12..355V}, \cite{10.2307/2289860} to stabilize the $\mSigma'$ in \eqref{mean_prime_sigma_prime} against singularities 
by introducing two hyperparameters that dictate the relative strength of the dimension variances and the off-diagonal covariance interactions as,
\begin{equation}
\mSigma'_{s} = \mSigma' + \alpha_1\sigma_1\rmI + \alpha_2 \sigma_2(\1 -  \rmI),
\label{cov_shrinking}
\end{equation}
where $\sigma_1$ is the average diagonal variance and $\sigma_2$ is the average off-diagonal covariance of $\mSigma'$. 

\subsubsection{Sampling the novel class}
\label{Sampling_the_novel_class}
With $\rvmu', \mSigma'_s$ now formulated, we have both the mean and covariance to represent the ground truth distribution of the novel class associated with the support point $\rvxtilda$. Hence we can sample $n$ points from this extrapolated distribution and append them to the existing support set,
\begin{equation}
    \sD_y = \{(\rvx,y) | \rvx \sim \gN(\rvmu', \mSigma'_s) \}
\label{sample_novel_points}
\end{equation}
where $y$ denotes the class of $\rvxtilda$. All the above steps discussed in subsections of \ref{sec:algorithm} are repeated for every point $\rvxtilda$ in the support set $\sS$, which has $N$ classes with $K$ points in each class constituting an $N$way-$K$shot classification task $\gT$. Hence the total dataset $\sD$ after augmentation can be written as, 
\begin{equation}
    \sD = \cup\{\sD_y \forall (\rvxtilda,y) \in \sS \} \cup \sS
\label{entire_dataset}
\end{equation}
$\sD$ can be used to construct a classifier like Logistic Regression, SVM etc. and the performance is reported as the average accuracy on the query set $\sQ$ in $\gT$.

\begin{algorithm}
\caption{GDC-- Generalized Distribution Calibration for Few-Shot Learning}\label{alg:cap}
\begin{algorithmic}
\Require Base class features $\rmX_i\in \sR^{d}, i \in \sC_b$
\Require Support set features $\sS=\{(\rvxtilda_j, y_j)\}_{j=1}^{N\times K} : \rvxtilda_j \in \sR^d, y_j \in \sC_n$
\For {$(\rvxtilda_j, y_j) \in \sS$}
    \State Gaussianize $\rvxtilda$ with \eqref{TukeyorYJ}
    \State Find the nearest $k$ base classes using $d_i$, $\sS_k$ (\eqref{find_di}, \ref{find_sk})
    \State Calculate weights $w_i$ for each base class $i \in \sS_k$ (\eqref{weights})
    \State Use $w_i$ to find the weighted random variable $\rmX'$ (\eqref{weighted-rv})
    \State Calculate $\rvmu', \mSigma'$ (\eqref{mean_prime_sigma_prime})
    \State Shrink $\mSigma'$ to get $\mSigma'_s$ (\eqref{cov_shrinking})
    \State Sample features for class $y_j$ using $\{(\rvx,y_j) | \rvx \sim \gN(\rvmu', \mSigma'_s) \}$ (\eqref{sample_novel_points})
\EndFor
\State Construct $\sD$ by appending all sampled features to the support set $\sS$  (\eqref{entire_dataset})
\State Train a logistic regression classifier\footnotemark on $\sD$ 
\State Report the accuracy on query set $\sQ=\{(\rvxtilda_j, y_j)\}_{j=N\times K+1}^{N\times K + N \times q}$
\label{our_algorithm}
\end{algorithmic}
\end{algorithm}
\footnotetext{Details of Logistic Regression is in Supplementary section \ref{Details_of_Logistic_Regression}}

\section{Experiments}
In this section we compare the performance of GDC with other states-of-the-art, show that our sampled points are closer to query data than DC, give theoretical insights with empirical results on the generalization improvement of GDC, and perform an ablation study to show the effectiveness of each component in our method.

\subsection{Implementation Details}
\subsubsection{Datasets}
We compare our proposed method with existing states-of-the-art on \textit{mini}Imagenet \cite{DBLP:conf/iclr/RaviL17},  CUB \cite{WelinderEtal2010} and  Stanford Dogs \cite{KhoslaYaoJayadevaprakashFeiFei_FGVC2011}. We follow the same train/test/validation splits as done in previous works \cite{DBLP:conf/iclr/RaviL17,DBLP:journals/corr/abs-1904-04232,chen2021multilevel}\footnote{More details on these datasets are discussed in Supplementary section \ref{dataset-information}}. To show that our method gives superior performance when the base classes are dissimilar to the novel classes, we evaluate our method on  Cross Domain dataset by training on tasks sampled from one distribution and evaluating on a different distribution. Specifically, we follow \cite{DBLP:conf/nips/PatacchiolaTCOS20} and show results on \textit{mini}Imagenet $\xrightarrow[]{}$ CUB, i.e. train split from miniImagenet and test/validation split from CUB. We also compare our method with DC \cite{DBLP:conf/iclr/YangLX21} on a meta-tieredImagenet dataset of the 34 broad categories from tieredImagenet, split into 20 base, 8 novel and 6 validation classes, as laid out in \cite{ren2018metalearning}. We assured that there is a high dissimilarity between the base and novel/validation classes in this meta-tieredImagenet (as noted in Table \ref{meta-dataset-split} of Supplementary section \ref{dataset-information}).

\subsubsection{Feature Extractor}
We used a WRN-28-10 \cite{DBLP:journals/corr/ZagoruykoK16} feature extractor trained using S2M2 Method \cite{mangla2020charting} for miniImagenet, CUB, Stanford Dogs and meta-tieredImagenet experiments. For our cross domain results on miniImagenet $\xrightarrow[]{}$ CUB, we used a Conv-4 backbone to compare our results with other states-of-the-art as done in \cite{DBLP:conf/nips/PatacchiolaTCOS20}\footnote{Training details of each feature extractor can be found in Supplementary section \ref{feature-extractor-backbone}}.

The extracted features were 640 dimensional for miniImagenet and CUB which had 600 and 44 points in each base class. Feature dimensions were 1600 for miniImagenet $\xrightarrow[]{}$ CUB experiments with 600 points in each base class. Note that such mismatch between the feature dimensions and the number of datapoints leads to singularities when trying to estimate the covariance matrix. Our shrinkage method takes care of these pathological situations. To show that our proposed algorithm works equally well for non-singular cases, we extracted 64 dimensional representation for Stanford Dogs and meta-tieredImagenet which had 148 and 12950 points in each of their base classes.

To demonstrate that our gaussianization approach (Section \ref{GaussianizationData}) generalizes, our features in miniImagenet and CUB come from a penultimate \texttt{relu} layer hence are all non negative. For Stanford Dogs and meta-tieredImagenet, we project 640 dimensional features from WRN-28-10 to 64 dimensions in the penultimate layer and hence the features span in both positive and negative regions.


\begin{table}[t]
\caption{Comparing the results of our proposed algorithm GDC on Stanford Dogs and CUB with 95\% confidence intervals. Best results highlighted in bold.}
\label{tab:CUBSF-results}
\begin{center}
\resizebox{\textwidth}{!}{
\begin{tabular}{l*{5}{c}r}
    \hline
        & \multicolumn{2}{c}{Stanford Dogs}
        & \multicolumn{2}{c}{CUB} \\
    Methods & 5way-1shot & 5way-5shot & 5way-1shot & 5way-5shot \\
    \hline
    RelationNet \cite{DBLP:conf/cvpr/SungYZXTH18} & 43.33 ± 0.42 & 55.23 ± 0.41 & 62.45 ± 0.98 & 76.11 ± 0.69\\
    adaCNN \cite{pmlr-v80-munkhdalai18a} & 41.87 ± 0.42 & 53.93 ± 0.44 & 56.57 ± 0.47 & 61.21 ± 0.42\\
    PCM \cite{Wei_2019} & 28.78 ± 2.33 & 46.92 ± 2.00 & 42.10 ± 1.96 & 62.48 ± 1.21\\
    CovaMNet \cite{Li_Xu_Huo_Wang_Gao_Luo_2019} & 49.10 ± 0.76 & 63.04 ± 0.65 & 60.58 ± 0.69 & 74.24 ± 0.68\\
    DN4 \cite{8953758} & 45.41 ± 0.76 & 63.51 ± 0.62 & 52.79 ± 0.86 & 81.45 ± 0.70\\
    PABN+cpt \cite{Huang_2021} & 45.65 ± 0.71 & 61.24 ± 0.62 & 63.56 ± 0.79 & 75.35 ± 0.58\\
    LRPABN+cpt \cite{Huang_2021} & 45.72 ± 0.75 & 60.94 ± 0.66 & 63.63 ± 0.77 & 76.06 ± 0.58\\
    MML \cite{chen2021multilevel} & 59.05 ± 0.68 & 75.59 ± 0.51 & 63.86 ± 0.67 & 80.73  ± 0.46 \\
    DC \cite{DBLP:conf/iclr/YangLX21} & - & - & 79.56 ± 0.87 & 90.67 ± 0.35 \\
    GDC (Ours) & \textbf{65.35 ± 0.61} & \textbf{80.56 ± 0.45} & \textbf{84.57 ± 0.48} & \textbf{93.46 ± 0.25} \\
    \hline
   \end{tabular}
}
\end{center}
\end{table}

\begin{table}[ht!]
\caption{Comparing the results of our proposed algorithm GDC on miniImagenet and Cross Domain miniImagenet $\xrightarrow[]{}$ CUB tasks with 95\% confidence intervals. Best results highlighted in bold.}
\label{tab:MI-MI-CUB-results}
\begin{center}
\resizebox{\textwidth}{!}{
\begin{tabular}{l*{5}{c}r}
    \hline
        & \multicolumn{2}{c}{miniImagenet}
        & \multicolumn{2}{c}{miniImagenet $\xrightarrow[]{}$ CUB} \\
    Methods & 5way-1shot & 5way-5shot & 5way-1shot & 5way-5shot \\
    \hline
        MatchingNet \cite{DBLP:journals/corr/VinyalsBLKW16} & 43.56 ± 0.84 & 55.31 ± 0.73 & 36.98 ± 0.06 & 50.72 ± 0.36 \\
        ProtoNet \cite{DBLP:conf/nips/SnellSZ17} & 54.16 ± 0.82 & 73.68 ± 0.65 & 33.27 ± 1.09 & 52.16 ± 0.17 \\
        MAML \cite{pmlr-v70-finn17a} & 48.70 ± 1.84 & 63.11 ± 0.92 & 34.01 ± 1.25 & 48.83 ± 0.62 \\
        RelationNet \cite{DBLP:conf/cvpr/SungYZXTH18} & 50.44 ± 0.82 & 65.32 ± 0.70 & 37.13 ± 0.20 & 51.76 ± 1.48 \\
        Baseline++ \cite{DBLP:journals/corr/abs-1904-04232} & 51.87 ± 0.77 & 75.68 ± 0.63 & 39.19 ± 0.12 & \textbf{57.31 ± 0.11} \\
        DKT \cite{DBLP:conf/nips/PatacchiolaTCOS20} & 49.73 ± 0.07 & 64.00 ± 0.09 & 40.22 ± 0.54 & 56.40 ± 1.34 \\
        E3BM \cite{liu2020ensemble}  & 63.80 ± 0.40 & 80.29 ± 0.25 & - & - \\
        Negative-Cosine \cite{liu2020negative} & 62.33 ± 0.82 & 80.94 ± 0.59 & - & - \\
        Meta Variance Transfer \cite{pmlr-v119-park20b} & - & 67.67 ± 0.70  & - & - \\
        DC \cite{DBLP:conf/iclr/YangLX21} & 68.57 ± 0.55 & 82.88 ± 0.42 & 35.08 ± 0.55 & 50.81 ± 0.43 \\
        GDC (ours) & \textbf{73.00 ± 0.50} & \textbf{87.22 ± 0.33} & \textbf{41.08 ± 0.53} & 54.69 ± 0.41 \\
    \hline
   \end{tabular}
    }
\end{center}
\end{table}

\subsubsection{Hyperparameter Search}
To circumvent combinatorial explosion from grid search of a larger number of hyperparameters, we used optuna \cite{optuna_2019} library for tuning our hyperparameters $\beta, m, k, \alpha_1, \alpha_2, n$. All tunings are 1000 trials long with a \texttt{Median Pruner}. Engineering details on our optuna setting and picking the best hyperparameters can be found in Supplementary section \ref{Hyperparameter_Search_Methodology}.

\subsection{Comparison with states-of-the-art}
Tables \ref{tab:CUBSF-results} and \ref{tab:MI-MI-CUB-results} summarize the performance of our proposed GDC with existing states-of-the-arts. We report the average classification accuracy of 5000 random tasks $\gT$ sampled from the novel classes $\sC_n$ along with their 95\% confidence intervals\footnote{The exact value of each hyperparameter that give the reported accuracy along with other candidate hyperparameters are given in Supplementary section \ref{Tuned_Hyperparameters}}.

We observe that a logistic regression constructed on our sampled data achieves consistent performance improvements of 3\% (CUB 5way-5shot)
to 5\% (CUB 5way-1shot) compared to the 2nd best method. We also achieve 1\% improvement on challenging cross domain 5way-1shot classification of \textit{mini}Imagenet $\xrightarrow[]{}$ CUB. 
Finally, for the case when base classes and novel classes share very little similarity (as is the case for the meta-tieredImagenet), we outperform DC by more than 9\% for 5way-5shot task as shown in Table \ref{tab:meta-results}.

Note that these accuracy improvements are derived from a simple statistical model that leverages just 3 additional hyperparameters compared to the 2nd best method of DC \cite{DBLP:conf/iclr/YangLX21}. 
We also do not make any assumptions about the feature values, the feature dimensions, and show these improvements across different cases of singular (miniImagenet, CUB, miniImagenet $\xrightarrow[]{}$ CUB) and non-singular covariances (Stanford Dogs, meta-tieredImagenet).

\begin{table}[t]
\caption{Comparing the results of our proposed algorithm GDC on meta-tieredImagenet 5way-1shot and 5way-5shot tasks with 95\% confidence intervals. The results below are reported for 5000 tasks sampled from the novel classes.}
\begin{center}
\begin{tabular}{l*{5}{c}r}
    \hline
        & \multicolumn{2}{c}{meta-tieredImagenet} \\
    Methods & 5way-1shot & 5way-5shot \\
    \hline
    No Method (baseline) & 20.00 ± 0.00 & 20.00 ± 0.00 \\
    DC \cite{DBLP:conf/iclr/YangLX21} & 39.54 ± 0.64 & 47.11 ± 0.78 \\
    GDC (Ours) & \textbf{43.51 ± 0.50} & \textbf{56.79 ± 0.64} \\
    \hline    
   \end{tabular}
\label{tab:meta-results}
\end{center}
\end{table}

\begin{figure}[ht!]
    \centering
    \includegraphics[scale=0.18]{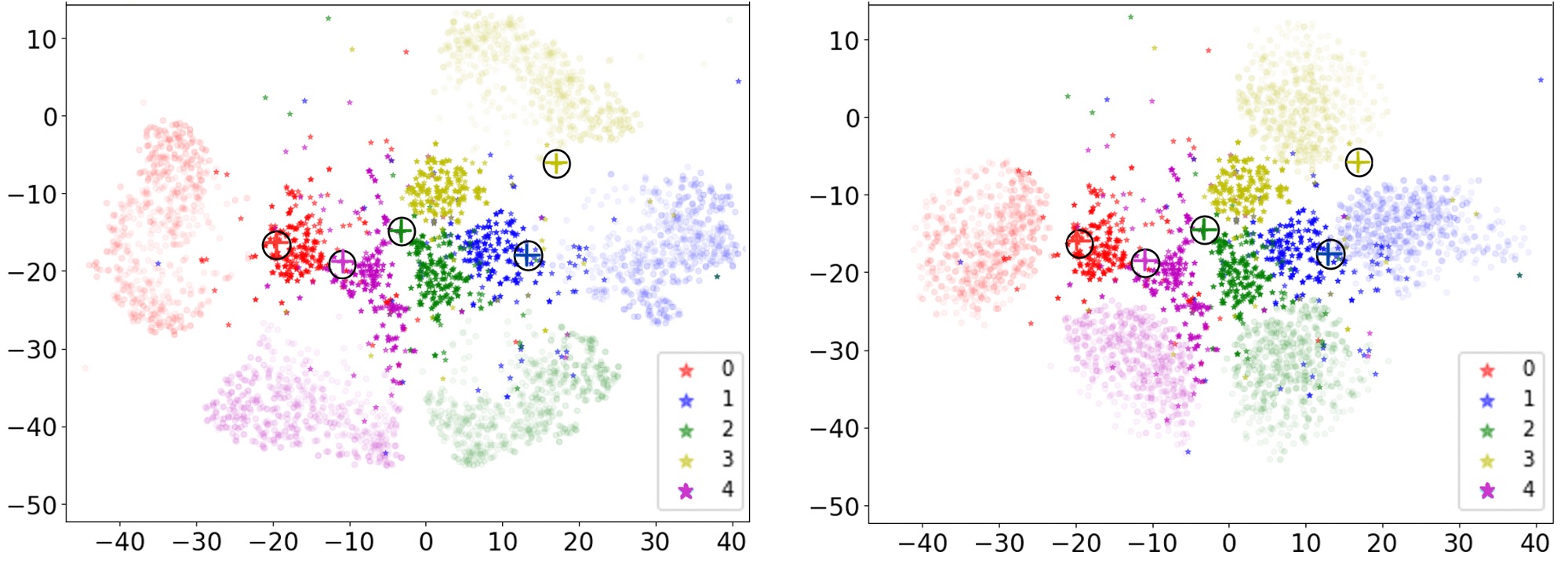}
    \caption{t-SNE visualization for 5 randomly sampled novel classes for DC (left) and our method GDC (right) in miniImagenet dataset. The support points are indicated with a '+' sign within black circles, sampled points are semi-transparent indicated with a 'o' sign, and the query points are opaque denoted by '*'. Note that our sampled points are on average closer to the query points.}
    \label{fig:tsne_1}
\end{figure}

\subsection{Visualization of Sampled Points}
In Figure \ref{fig:tsne_1} we compare the t-SNE representation \cite{JMLR:v9:vandermaaten08a} of sampled points for both DC \cite{DBLP:conf/iclr/YangLX21} and our method GDC to visualize whether our sampled points are closer to the ground truth as indicated quantitatively from results in Tables \ref{tab:CUBSF-results}, \ref{tab:MI-MI-CUB-results}, \ref{tab:meta-results}. We can see that our method produces clusters of sampled points which are more compact and overlap with more query points than the sampled points of DC. We analyse this in the next section \ref{why_dcplus_better}\footnote{For more visualizations refer to Supplementary section \ref{additional_tsne_visualization}}

\begin{figure}[t]
    \centering
    \includegraphics[scale=0.35]{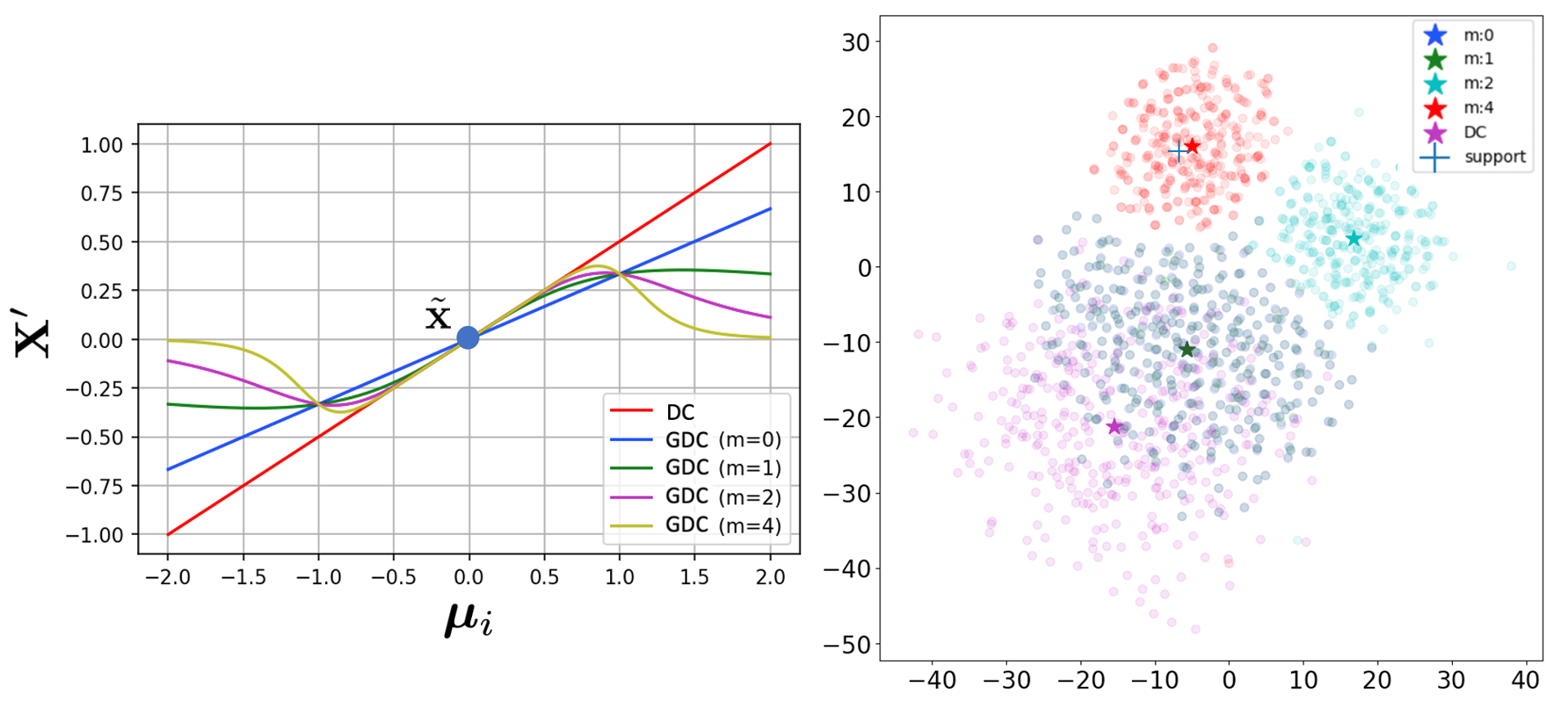}
    \caption{\textbf{Left}: Novel class estimate $\rmX'$ as a function of $\rvmu_i$ for a 1 dimensional example. $\rvxtilda$ is at origin. Note that as base class $i$ moves away from $\rvxtilda$, the error in DC's estimation of $\rmX'$ accumulates. With $m=4$, GDC still produces $\rmX'$ close to $\rvxtilda$.
    \textbf{Right}: t-SNE plots of sampled points along with their mean (denoted by '*') confirming analysis in \textbf{Left} figure. Note that the sampled points of $m=4$ overlaps with $\rvxtilda$. Sampled clusters of $m=0,1$ overlap with each other.}
    \label{fig:why_dcplus_higher}
\end{figure}

\subsection{Why should GDC generalize better than DC?}
\label{why_dcplus_better}
To motivate the generalization and performance improvements of GDC, we start with a comparison of the weighting scheme between DC and GDC. Figure \ref{fig:why_dcplus_higher} (left) shows the estimated position of $\rmX'$ as the closest base class moves away from $\rvxtilda$ for the simple case of one base class. 
The unweighted random variable $\rmX'$ in DC can be written as $\rmX' = (\rvxtilda + \sum_{i \in \sS_k}\rmX_i)/(1+k) = (\rvxtilda + \rmX_1) / 2$, for $k$=1.


Hence as $\rvmu_i = E(\rmX_1)$ moves away from $\rvxtilda$, $\rmX'$ moves further apart in DC as seen in Figure \ref{fig:why_dcplus_higher}. Thus even though the method borrows stable values of mean and covariance from the base class in question, the values themselves are far from the support point, which acts as a proxy location for the query points during training. In contrast, with a higher $m$ in GDC, the weight $w_i$ assigned to base class $i$ drops polynomially with distance from $\rvxtilda$ and $\rmX'$ saturates at $\rvxtilda$ when the base class is far from $\rvxtilda$. Hence our sampled points should be much closer to $\rvxtilda$ giving us better generalization than DC.

To confirm this hypothesis, we visualize the sampled points for $k=1$ in miniImagenet $\xrightarrow[]{}$ CUB experiment in Figure \ref{fig:why_dcplus_higher}. We chose this dataset to also show how a higher $m$ can help offset noise from base classes by minimizing their $w_i$ in estimating the distribution $\rmX'$ when they are not similar to the novel class of $\rvxtilda$. We can see that points produced by DC and $m=0,1$ are far from the support point $\rvxtilda$ and overlap. As $m$ increases, the cluster moves closer to $\rvxtilda$ where at $m=4$ the cluster overlaps with $\rvxtilda$. Our peformance improvement of 5\% on 5way-1shot and 4\% on 5way-5shot tasks of Cross Domain miniImagenet $\xrightarrow[]{}$ CUB experiments compared to DC further confirms this hypothesis.

This same trend can also be seen in the results of meta-tieredImagenet in Table \ref{tab:meta-results} where GDC converges on a high value of $m$ ($m= 2.25$ in 5way-1shot and $m=2$ in 5way-5shot) giving weights to base classes that are 10 to 100 smaller than miniImagenet, and 100 times smaller than DC. We further analyse this case and visualize the histogram of these weights comparing them with DC, and weights from miniImagenet experiments in Supplementary section \ref{meta-dataset-appendix}.

\begin{table}[t]
\caption{Ablation study of GDC on Stanford Dogs 5way-1shot task showing the change in accuracy (with 95\% confidence interval over 200 tasks) as each component of our model is added incrementally. The cumulative improvement in accuracy is a significant $4\%$ compared to the baseline.}
\label{case-study-table}
\begin{center}
\resizebox{\textwidth}{!}{
\begin{tabular}{p{4cm}p{3cm}p{3cm}{r}}
    \hline
    & Fixed & Tuned & \\
    Step & Hyperparameter & Hyperparameter & 5way-1shot \\
    \hline
    Gaussianization & Baseline, only hyperparameter is $\beta$ & $\beta$=1 & 60.44 ± 0.98\% \\
    Top-k  Selection & $\alpha_1$=0, $\alpha_2$=0, $m$=0 & $k$=1 & 61.27 ± 0.97\% ($\uparrow$0.8\%) \\
    Distance-weighted random variable & $\alpha_1$=0, $\alpha_2$=0, $m$=1 & $k$=1 &   61.98 ± 0.98\% ($\uparrow$0.7\%) \\
    Shrinkage (with $\alpha_1$) & $\alpha_2$=0, $m$=1 & $k$=4, $\alpha_1$=100 & 63.46 ± 0.95\% ($\uparrow$1.5\%)\\
    Shrinkage (with $\alpha_1$ and $\alpha_2$) & $m$=1 & $k$=12,$\alpha_1$=400, $\alpha_2$=200$\alpha_1$ & 63.92 ± 0.97\% ($\uparrow$0.4\%) \\
    All parameters tuned simultaneously & None & $k$=10,$\alpha_1$=400, $\alpha_2$=100$\alpha_1$,$m$=1.5 & 64.33 ± 0.99\% ($\uparrow$0.4\%)\\
    \hline    
   \end{tabular}
}
\end{center}
\end{table}

\subsection{Ablation Study}
We ran a 5way-1shot ablation study on the Stanford Dogs dataset to estimate the effect of each hyperparameter of our method. For each experiment, the hyperparameter range were defined as,
\begin{flalign*}
    m \in & [low=0, high=3, step=0.25] & k \in & [low=1, high=20, step=1] \\
    \alpha_1 \in & [low=0, high=600, step=100] & \alpha_2 \in & [low=0 \times \alpha_1, high=600 \times \alpha_1, step=100]
\end{flalign*}    

We fixed $n=750$, and the total number of trials in our \texttt{optuna} hyperparameter search to 100 so that the accuracy improvements with the addition of each hyperparameter were independent of the search time. For each trial, the number of tasks $\gT$ was also fixed to 200.

Table \ref{case-study-table} summarizes the results of our study and shows that each component of GDC is important in giving improvements over state-of-the-art. For the Gaussianization step, the final logistic regression model was trained on one data point from each novel class in a 5way classification task. Notice that $\beta=1$ in our study implies  that raw features from 64 dimensional linear layer give best accuracy.
The results for $m=1$ show that introducing weighted random variables to model novel class distributions helps even without tuning $m$. Further control on decaying $w_i$ as a function of $d_i$ as discussed in Section \ref{why_dcplus_better} with a tuned $m$ helps in achieving state-of-the-art.

\begin{figure}[t]
    \centering
    \includegraphics[scale=0.37]{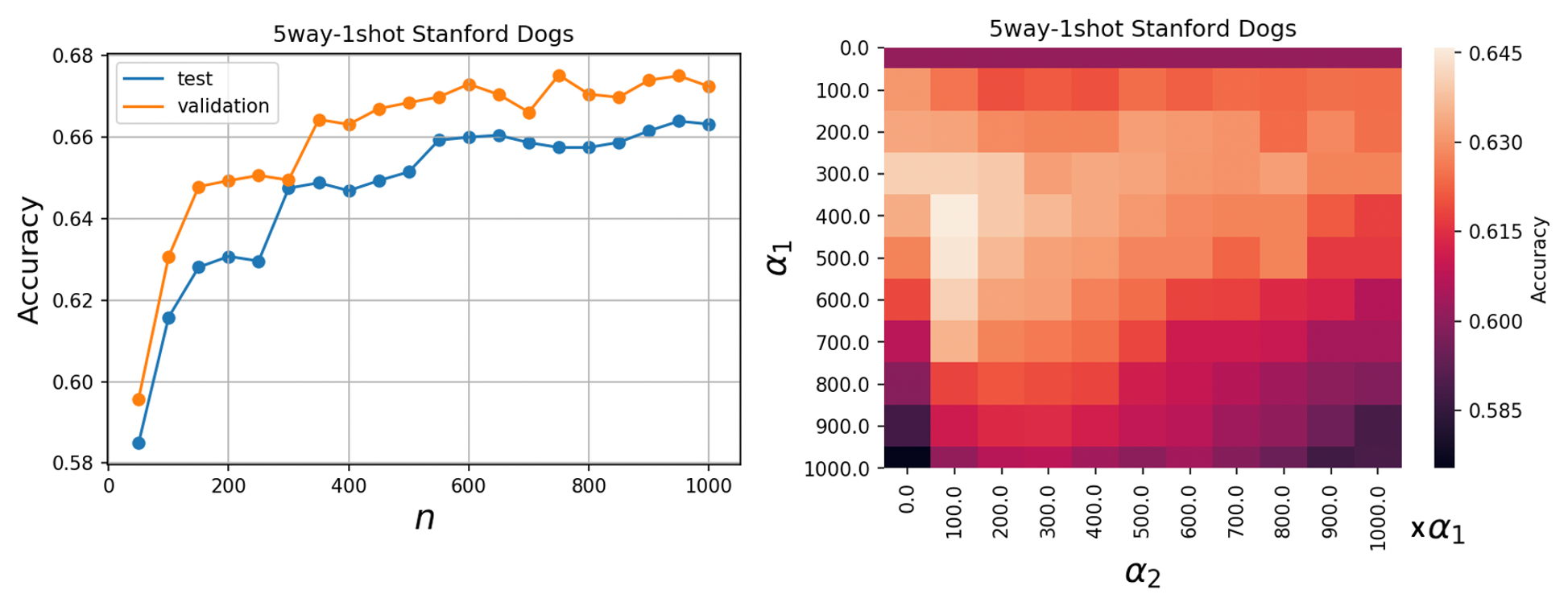}
    \caption{\textbf{Left}: Accuracy v/s $n$ the number of sampled features for validation and novel classes in 5way-1shot Stanford Dogs dataset.
    \textbf{Right}: Accuracy v/s $\alpha_1, \alpha_2$ in 5way-1shot Stanford Dogs dataset. The accuracy for each point for both $n$ and $\alpha_1, \alpha_2$ experiment is an average over 200 tasks.}
    \label{fig:acc_vs_n_alphas}
\end{figure}

\subsubsection{Effect of $k$}
We investigated whether the weight decay parameter $m$ can remove the need for the hyperparameter $k$, since increasing $m$ reduces the weight $w_i$ of a base class that is farther from a given novel point. In Table \ref{case-study-table} we see that the best $k$ equals 10 for the case when all hyperparameters are tuned together. This indicates that $k$ and $m$ work together to estimate novel class statistics (otherwise the best value would be $k=30=|\sC_b|$, number of base classes in Stanford Dogs dataset). To see how much the introduction of $k$ is helping the accuracy, we tuned all hyperparameters in Stanford Dogs 5way-1shot task and find that without $k$, the accuracy was 63.46\% with $m=1.5, \alpha_1=600, \alpha_2=0$. This was 0.9\% lower than the best result with all $k,\alpha_1,\alpha_2,m$ tuned in the last row of Table \ref{case-study-table}. 

\subsubsection{Effect of $n, \alpha_1, \alpha_2$}
The value of $n$, the number of sampled points, was fixed to 750 in all our experiments. Figure \ref{fig:acc_vs_n_alphas} shows the trend of accuracy v/s $n$ in our ablation study on Stanford Dogs 5way-1shot task. We can see that the accuracy keeps increasing with slight noise until $n=750$ after which it saturates. It indicates that the sampling more points after a certain number does not cover additional query points in this specific feature space.

From Table \ref{case-study-table} we see that $\alpha_1, \alpha_2$ play a key role in improving accuracy. In Figure \ref{fig:acc_vs_n_alphas} the heatmap of accuracy versus $\alpha_1$ and $\alpha_2$ shows that accuracy increases with increasing $\alpha_1$, $\alpha_2$ and then starts dropping beyond $\alpha_1>200$. This could be due to overlapping clusters of sampled points with a higher covariance, where the decision boundaries start overlapping. 

\section{Conclusion}
In this work, we have proposed a principled approach to estimate novel class distributions by formulating a similarity-based weighted random variable of closest base classes. We showed that incorporating statistical techniques of covariance shrinkage and Gaussianization not only generalize our method (GDC) to arbitrary pretrained feature extractors, but also increase the accuracy over state-of-the-art significantly. 
Our experiments 
effectively demonstrate cumulative performance improvements of 1\% to 9\% over DC including challenging cross domain tasks. Exploring the trade-off between more hyperparameters and accuracy along with generalizations to different tasks like non-Gaussian distributions can be productive avenues to pursue.

\bibliographystyle{splncs04}
\bibliography{mybibliography}








\newpage

\section*{Supplementary}
\appendix



\section{Effect of Distances}
\label{Effect_of_Distances}

For normally distributed features, the probability of a novel point $\rvxtilda$ being in class $i$ with distribution $\gN(\rvmu_i, \mSigma_i)$ can be written quite generally as,
\begin{equation}
p(class=i|\rvxtilda) \approx p(i) \frac{1}{\sqrt{\mid\mSigma_{i}\mid}} \exp{\frac{-1}{2}(\rvxtilda-\rvmu_i)^T \Sigma^{-1}_{i} (\rvxtilda - \rvmu_i) },
\end{equation}
Taking $\ln$ on both sides of this probability, and assuming that each base class is equally probable in priors $p(i)$, we get a natural distance between $\rvxtilda$ and $\rvmu_i$ as:
\begin{equation}
d_i = \ln{\mid \mSigma_{i}\mid } + (\rvxtilda-\rvmu_i)^T \mSigma^{-1}_{i} (\rvxtilda - \rvmu_i)
\label{md_with_log}
\end{equation}

\eqref{md_with_log} is the squared Mahalanobis distance between $\rvxtilda$ and $\rvmu_i$ along with an added $\ln$ term. 
Replacing $\mSigma^{-1}_{i}=\rmI$ for all $i$, our $d_i$ reduces to squared Euclidean distance,
\begin{equation}
d_i = (\rvxtilda-\rvmu_i)^T (\rvxtilda - \rvmu_i)
\label{euclidean}
\end{equation}
The above formulations assume that both $\rvxtilda, \rvmu_i$ come from the same distribution. Keeping this in mind, we also derive a metric called squared $\delta$ distance with $\delta$ as a hyperparameter in Formulation 1 (Section \ref{General_Distance}) as,
\begin{equation}
    d_i = (\rmI\rvxtilda - \delta\rmI\rvmu_i)^{T}(\rmI\rvxtilda - \delta\rmI\rvmu_i)
\label{general_distance_main_text}
\end{equation}


\begin{table}[ht!]
\caption{Accuracy with different distance measures in Stanford Dogs 5way-1shot task. Note that $m=1$ in all distances except the scaled Euclidean distance in the last row.}
\label{distances-vs-accuracy}
\begin{center}
\begin{tabular}{l*{2}{c}r}
    \hline
        Distance ($d_i$) & 5way-1shot \\
    \hline
    Squared Mahalanobis with log, \eqref{md_with_log}     &   63.95\% \\
    Squared Euclidean, \eqref{euclidean}                &   64.33\% \\
    Squared $\delta$ distance, \eqref{general_distance_main_text}  &   64.57\% \\
    Scaled Euclidean $m=1.5$          &   64.58\% \\
    \hline
   \end{tabular}
\end{center}
\end{table}

Experimenting with these different distance metrics, in Table \ref{distances-vs-accuracy}, we observe no advantage of using Mahalanobis distance or squared $\delta$ distance over a simpler Euclidean distance when defining $d_i$ in Section \ref{Proposed_Random_Variable}. We also see that the scaled Euclidean with $m$ as a hyperparameter gives improved accuracy compared to Euclidean with $m=1$. Hence, we propose using our scaled Euclidean distance (\eqref{weights} in main text) instead of a searching for multiple distance metrics in the feature space to improve accuracy. 

\subsection{Formulation 1: General Distance: Squared $\delta$}
\label{General_Distance}
Consider two points $\rvx_1, \rvx_2 \in \sR^d$ sampled from different multivariate distributions $\gN(\rvmu_1, \mSigma_1), \gN(\rvmu_1, \mSigma_1), \rvmu_1, \rvmu_2 \in \sR^d, \mSigma_1, \mSigma_2 \in \sR^{d\times d}$. Multiple distance measures can be constructed between the populations, e.g., the Bhattacharya distance \cite{bhattacharya1943}. Here, we consider a general distance of the form
\begin{flalign}
    d = (\rmL^{-1}\rvx_1 - \rmM^{-1}\rvx_2)^{T}(\rmL^{-1}\rvx_1 - \rmM^{-1}\rvx_2) \label{general_distance_appendix}\\
    where, \;\;\rmL \rmL^T = \mSigma_1, \;\;\;\; \rmM \rmM^T = \mSigma_2 \nonumber
\end{flalign}
Here $\rmL, \rmM \in \sR^{d\times d}$ are the cholesky decompositions of $\mSigma_1, \mSigma_2$ respectively. If $\mSigma_1=\mSigma_2$, equation \ref{general_distance_appendix} reduces to squared Mahalanobis distance.
If $\rvx_1$ is the support point $\rvxtilda$ and $\rvx_2$ is the mean of base class $\rmX_i$, then $\mSigma_2=\mSigma_i=$covariance of the base class $i$. The covariance of the distribution of $\rvxtilda$ is unknown since we are trying to estimate it. Assuming $\mSigma_1 = \psi\rmI$, i.e. a diagonal covariance, we get,
\begin{flalign}
    d_i =& (\rmL^{-1}\rvxtilda - \rmM^{-1}\rvmu_i)^{T}(\rmL^{-1}\rvxtilda - \rmM^{-1}\rvmu_i)\nonumber\\
        =& (\sqrt{\psi}\rmI\rvxtilda - \rmM^{-1}\rvmu_i)^{T}(\sqrt{\psi}\rmI\rvxtilda - \rmM^{-1}\rvmu_i)\nonumber\\
    & where, \;\;\rmL \rmL^T = \psi\rmI, \;\;\;\; \rmM \rmM^T = \mSigma_i \nonumber
\end{flalign}
From our experiments we observed that the off-diagonal covariances in the base classes of \textit{mini}Imagenet, CUB and Stanford Dogs, were at least 2 orders of magnitudes smaller than the diagonal variance. Hence assuming $\mSigma_i=\sigma\rmI$, i.e. a diagonal matrix with constant variance, we get,
\begin{flalign}
    d_i =& (\sqrt{\psi}\rmI\rvxtilda - \sqrt{\sigma}\rmI\rvmu_i)^{T}(\sqrt{\psi}\rmI\rvxtilda - \sqrt{\sigma}\rmI\rvmu_i)\nonumber\\
        =& \psi(\rmI\rvxtilda - \frac{\sqrt{\sigma}}{\sqrt{\psi}}\rmI\rvmu_i)^{T}(\rmI\rvxtilda - \frac{\sqrt{\sigma}}{\sqrt{\psi}}\rmI\rvmu_i)\nonumber\\
        =& \psi(\rmI\rvxtilda - \delta\rmI\rvmu_i)^{T}(\rmI\rvxtilda - \delta\rmI\rvmu_i)\;\;, where \;\;\delta =  \frac{\sqrt{\sigma}}{\sqrt{\psi}}
\end{flalign}

Since $\psi$ is common to all base classes $\rmX_i$, it does not affect our closest base class calculation. Hence dropping we $\psi$ we get a general distance as,
\begin{flalign}
    d_i = (\rmI\rvxtilda - \delta\rmI\rvmu_i)^{T}(\rmI\rvxtilda - \delta\rmI\rvmu_i)
\end{flalign}
where $\delta$ is a hyperparameter that can be optimized for accuracy.

\section{Details of Logistic Regression}
\label{Details_of_Logistic_Regression}
We performed Logistic Regression using \texttt{torch} library. For all datasets– \textit{mini}Imagenet, CUB, StanfordDogs and for all experiments, the following hyperparameters were used,\\
\begin{flalign*}
\text{batch size} &= 1024, \\
\text{epochs} &=200, \\
\text{learning rate} &=0.08, \\
\text{optimizer} &= \text{Stochastic Gradient Descent} \;(\texttt{torch.optim.SGD}), \\
\text{scheduler} &= \text{None}, \\
\text{Loss Function} &= \text{Cross Entropy} \;(\texttt{torch.nn.CrossEntropyLoss}), \\
\text{Loss Regularization} &= \text{None}\\
\end{flalign*}

\section{Datasets}
\label{dataset-information}

\textbf{\textit{mini}Imagenet} \cite{DBLP:conf/iclr/RaviL17} is derived from the ILSVRC-12 dataset \cite{DBLP:journals/corr/RussakovskyDSKSMHKKBBF14}. It contains 100 different classes with 600 examples per class. The images sizes are $84\times 84 \times 3$. We follow the train, validation, test split of 64 base, 16 validation and 20 novel classes as done in previous work of \cite{DBLP:conf/iclr/RaviL17}.

\textbf{CUB} \cite{WelinderEtal2010} is a fine grained classification dataset consisting of different bird species. Each class has varying number of examples in this dataset so we take the minimum available number of 44 examples from each of 200 classes. The image sizes are again $84 \times 84 \times 3$. Following \cite{DBLP:journals/corr/abs-1904-04232} we split train, validation, test as 100 base, 50 validation and 50 novel classes respectively.

\textbf{Stanford Dogs} \cite{KhoslaYaoJayadevaprakashFeiFei_FGVC2011} is another fine grained classification dataset of dogs species derived from ILSVRC-12 dataset. Here again each class has varying number of points so we take 100 points from each of 120 classes. Following existing state-of-the-art results of \cite{chen2021multilevel}, we use train, validation and test splits of 70 base, 20 validation and 30 novel classes respectively.

\textbf{Cross Domain datasets}: To show that our method gives superior performance even when the base classes are dissimilar to the novel classes, we evaluate our proposed method by training on tasks sampled from one distribution and evaluating on a different distribution. Specifically, we follow \cite{DBLP:conf/nips/PatacchiolaTCOS20} and show results on \textit{mini}Imagenet $\xrightarrow[]{}$ CUB, i.e. train split from \textit{mini}Imagenet and test/val split from CUB. We also compare our method against DC \cite{DBLP:conf/iclr/YangLX21} on a \textbf{meta-tieredImagenet} of 34 broad categories from tieredImagenet, split into 20 base, 8 novel and 6 validation classes, as laid out in \cite{ren2018metalearning}. Note that there is a high dissimilarity between the base and novel/validation classes in this meta-tieredImagenet as seen in Table \ref{meta-dataset-split}.

\begin{table}[ht!]
\caption{34 broad categories of tieredImagenet dataset forming a meta-tieredImagenet. We can see that there is a high dissimilarity between the base and novel classes here. The only similar class between base and novel is ‘working dog’ and ‘hound, hound dog’.}
\begin{center}
\resizebox{\textwidth}{!}{
\begin{tabular}{llll}
\hline \\
\#Num & Base class                                     & Novel class                                     & Validation class                           \\ \hline
1     & protective covering, & obstruction, obstructor, & durables, durable goods,  \\
      & protective cover, protect   & obstructer, impediment & consumer durables \\
2     & Garment                                        & geological formation, formation                 & motor vehicle, automotive vehicle          \\
3     & building, edifice                              & solid                                           & machine                                    \\
4     & establishment                                  & substance                                       & furnishing                                 \\
5     & electronic equipment                           & vessel                                          & mechanism                                  \\
6     & game equipment                                 & aquatic vertebrate                              & sporting dog, gun dog                      \\
7     & Tool                                           & working dog                                     &                                            \\
8     & Craft                                          & insect                                          &                                            \\
9     & ungulate, hoofed mammal                        &                                                 &                                            \\
10    & musical instrument, instrument                 &                                                 &                                            \\
11    & Primate                                        &                                                 &                                            \\
12    & feline, felid                                  &                                                 &                                            \\
13    & hound, hound dog                               &                                                 &                                            \\
14    & Terrier                                        &                                                 &                                            \\
15    & snake, serpent, ophidian                       &                                                 &                                            \\
16    & Saurian                                        &                                                 &                                            \\
17    & passerine, passeriform bird                    &                                                 &                                            \\
18    & aquatic bird                                   &                                                 &                                            \\
19    & restraint, constraint                          &                                                 &                                            \\
20    & instrument                                     &                                                 &                      \\                     
\end{tabular}
}
\label{meta-dataset-split}
\end{center}
\end{table}

\newpage
\section{Feature Extractor Backbone}
\label{feature-extractor-backbone}
We used S2M2 Method \cite{mangla2020charting} to train a WRN-28-10 \cite{DBLP:journals/corr/ZagoruykoK16} feature extractor for \textit{mini}Imagenet, CUB, Stanford Dogs and meta-tieredImagenet. The backbone was first allowed to overfit for 400 epochs on all base classes to minimize the classification + self supervised rotation loss. Next the decision boundaries were smoothened using the validation classes with a cosine classifier on the feature extractor until the loss on validation classes stopped improving. For our cross domain results on \textit{mini}Imagenet $\xrightarrow[]{}$ CUB, we used identical setting as \cite{DBLP:conf/nips/PatacchiolaTCOS20} with a Conv-4 backbone.

\section{Hyperparameter Search Methodology}
\label{Hyperparameter_Search_Methodology}
We used \texttt{optuna} \cite{optuna_2019} library to tune our hyperparameters $\beta, m,k,\alpha_1,\alpha_2, n$. For all datasets (\textit{mini}Imagenet, CUB, Stanford Dogs, meta-tieredImagenet and \textit{mini}Imagenet $\xrightarrow[]{}$ CUB) the search space for $\beta, m, k, n$ was,
\begin{flalign*}
\beta \in [low=0,high=10,step=0.25], \; & m \in [low=0, high=3, step=0.25], \\
k \in [low=2, high=|\sC_b|, step=2], \; & n \in [low=100,high=1000,step=50]
\end{flalign*}

where $|\sC_b|$ denotes the number of base classes.

For \textit{mini}Imagenet and CUB we set
\begin{flalign*}
\alpha_1 \in [low=0,high=10000,step=1000], \alpha_2 \in \{0,0.1,1,10,100\}\times\alpha_1    
\end{flalign*}
whereas for Stanford Dogs, meta-tieredImagenet, and \textit{mini}Imagenet $\xrightarrow[]{}$ CUB,
\begin{flalign*}
\alpha_1 \in [low=0,high=1000,step=100], \alpha_2 \in [low=0, high=1000, step=100] \times \alpha_1
\end{flalign*}
Note that a larger range of $\alpha_2$ was needed in Stanford Dogs owing to the average off-diagonal covariance of $\mSigma'$ (\eqref{mean_prime_sigma_prime} in main text) being 2 orders of magnitude smaller than the average variance.

All hyperparameters were jointly tuned using \texttt{TPESampler} in \texttt{optuna}. For each hyperparameter setting sampled, we evaluated its average accuracy over 200 random tasks $\gT$ sampled from the validation classes $\sC_v$. During this phase, we pruned any hyperparameter setting which had less than median accuracy after 100 runs using \texttt{MedianPruner}. Once every hyperparameter setting was validated, we picked the top-3 candidates from this experiment and ran these specific hyperparameters for 5000 random tasks sampled from the novel classes, i.e. $\gT \sim \sC_n$. We report our accuracy and mean confidence of the best candidate in comparison with state-of-the-art in Tables \ref{tab:CUBSF-results}, \ref{tab:MI-MI-CUB-results} of main text. Accuracies of all 3 candidates can be found in Supplementary section \ref{Tuned_Hyperparameters}.

\newpage
\section{Tuned Hyperparameters}
\label{Tuned_Hyperparameters}
In Table \ref{candidate_accuracies}, we give the accuracy of each of the top 3 candidates from our \texttt{optuna} hyperparameter search. Note than the first row of every setting (5way-1shot or 5way-5shot) is the result we used for comparing to other methodologies in Tables \ref{tab:CUBSF-results}, \ref{tab:MI-MI-CUB-results} of the main text.

\begin{table}[ht!]
\caption{Accuracy for each of top 3 candidates during our \texttt{optuna} hyperparameter search. The accuracies are reported after evaluating on 5000 random tasks sampled from the novel classes, i.e. $\gT \sim \sC_n$}
\label{candidate_accuracies}
\begin{center}
\resizebox{\textwidth}{!}{
\begin{tabular}{l|lll}\hline
  Dataset & Setting & Accuracy & Hyperparameters \\ \hline
  \multirow{6}{*}{miniImagenet} & \multirow{3}{*}{5way-1shot} & 
  73.00 +- 0.50 & $m=1,k=8,\alpha_1=3000,\alpha_2=10\alpha_1, \beta=0.5, n=750$ \\
  & & 
  72.80 +- 0.43 & $m=1,k=8,\alpha_1=6000,\alpha_2=10\alpha_1, \beta=0.5, n=750$ \\
  & & 
  72.30 +- 0.51 & $m=1,k=8,\alpha_1=2000,\alpha_2=10\alpha_1, \beta=0.5, n=750$ \\ \cline{2-4}
  & \multirow{3}{*}{5way-5shot} & 
  87.22 +- 0.33 & $m=3,k=30,\alpha_1=9000,\alpha_2=10\alpha_1, \beta=0.5, n=750$ \\
  & & 
  86.91 +- 0.32 & $m=3,k=30,\alpha_1=8000,\alpha_2=10\alpha_1, \beta=0.5, n=750$ \\
  & & 
  86.82 +- 0.39 & $m=3,k=30,\alpha_1=7000,\alpha_2=10\alpha_1, \beta=0.5, n=750$ \\ \cline{1-4}
  \multirow{6}{*}{CUB} & \multirow{3}{*}{5way-1shot} & 
  84.57 +- 0.48 & $m=1,k=4,\alpha_1=8000,\alpha_2=10\alpha_1, \beta=0.5, n=750$ \\
  & & 
  84.49 +- 0.50 & $m=1,k=6,\alpha_1=10000,\alpha_2=10\alpha_1, \beta=0.5, n=750$ \\ 
  & & 
  84.22 +- 0.46 & $m=1,k=4,\alpha_1=9000,\alpha_2=10\alpha_1, \beta=0.5, n=750$ \\ \cline{2-4}
  & \multirow{3}{*}{5way-5shot} & 
  93.46 +- 0.25 & $m=2,k=4,\alpha_1=5000,\alpha_2=10\alpha_1, \beta=0.5, n=750$ \\
  & & 
  93.16 +- 0.21 & $m=1,k=10,\alpha_1=2000,\alpha_2=100\alpha_1, \beta=0.5, n=750$ \\
  & & 
  93.15 +- 0.26 & $m=1,k=16,\alpha_1=3000,\alpha_2=100\alpha_1, \beta=0.5, n=750$ \\ \cline{1-4}
  \multirow{6}{*}{StanfordDogs} & \multirow{3}{*}{5way-1shot} & 
  65.35 +- 0.61 & $m=1.5,k=10,\alpha_1=400,\alpha_2=100\alpha_1, \beta=1, n=750$ \\
  & & 
  65.12 +- 0.61 & $m=1.75,k=10,\alpha_1=500,\alpha_2=100\alpha_1, \beta=1, n=750$ \\
  & &
  64.91 +- 0.58 & $m=1.5,k=12,\alpha_1=500,\alpha_2=100\alpha_1, \beta=1, n=750$ \\ \cline{2-4}
  & \multirow{3}{*}{5way-5shot} & 
  80.56 +- 0.45 & $m=1.5,k=12,\alpha_1=300,\alpha_2=200\alpha_1, \beta=1, n=750$ \\ 
  & & 
  80.12 +- 0.46 & $m=1.25,k=12,\alpha_1=300,\alpha_2=400\alpha_1, \beta=1, n=750$ \\ 
  & & 
  80.01 +- 0.41 & $m=1.25,k=12,\alpha_1=100,\alpha_2=200\alpha_1, \beta=1, n=750$ \\ \cline{1-4}
    \multirow{6}{*}{miniImagenet $\xrightarrow[]{}$ CUB} & \multirow{3}{*}{5way-1shot} & 
  41.08 +- 0.53 & $m=0.5,k=64,\alpha_1=400,\alpha_2=100\alpha_1, \beta=0.5, n=750$ \\
  & & 
  39.11 +- 0.51 & $m=0.5,k=62,\alpha_1=500,\alpha_2=100\alpha_1, \beta=0.5, n=750$ \\
  & &
  39.02 +- 0.58 & $m=2.5,k=10,\alpha_1=400,\alpha_2=100\alpha_1, \beta=0.5, n=750$ \\ \cline{2-4}
  & \multirow{3}{*}{5way-5shot} & 
  54.69 ± 0.41 & $m=0.5,k=18,\alpha_1=100,\alpha_2=100\alpha_1, \beta=0.5, n=750$ \\ 
  & & 
  52.12 +- 0.45 & $m=0.5,k=12,\alpha_1=100,\alpha_2=100\alpha_1, \beta=0.5, n=750$ \\ 
  & & 
  49.14 +- 0.44 & $m=1.5,k=18,\alpha_1=100,\alpha_2=200\alpha_1, \beta=0.5, n=750$ \\ \cline{1-4}
    \multirow{6}{*}{meta-tieredImagenet} & \multirow{3}{*}{5way-1shot} & 
  43.51 ± 0.50 & $m=2.25,k=8,\alpha_1=100,\alpha_2=100\alpha_1, \beta=1, n=750$ \\
  & & 
  41.12 +- 0.61 & $m=2.0,k=10,\alpha_1=100,\alpha_2=100\alpha_1, \beta=1, n=750$ \\
  & &
  40.01 +- 0.58 & $m=1.5,k=12,\alpha_1=200,\alpha_2=100\alpha_1, \beta=1, n=750$ \\ \cline{2-4}
  & \multirow{3}{*}{5way-5shot} & 
  56.79 ± 0.64 & $m=2.0,k=10,\alpha_1=100,\alpha_2=100\alpha_1, \beta=1, n=750$ \\ 
  (tieredImagenet) & & 
  53.13 +- 0.66 & $m=2.25,k=12,\alpha_1=200,\alpha_2=100\alpha_1, \beta=1, n=750$ \\ 
  & & 
  50.01 +- 0.61 & $m=1.25,k=12,\alpha_1=100,\alpha_2=200\alpha_1, \beta=1, n=750$ \\ \cline{1-4}
\end{tabular}
}
\end{center}
\end{table}

\newpage
\section{Additional t-SNE Visualization}
\label{additional_tsne_visualization}
\begin{figure}[ht!]
    \centering
    \includegraphics[scale=0.18]{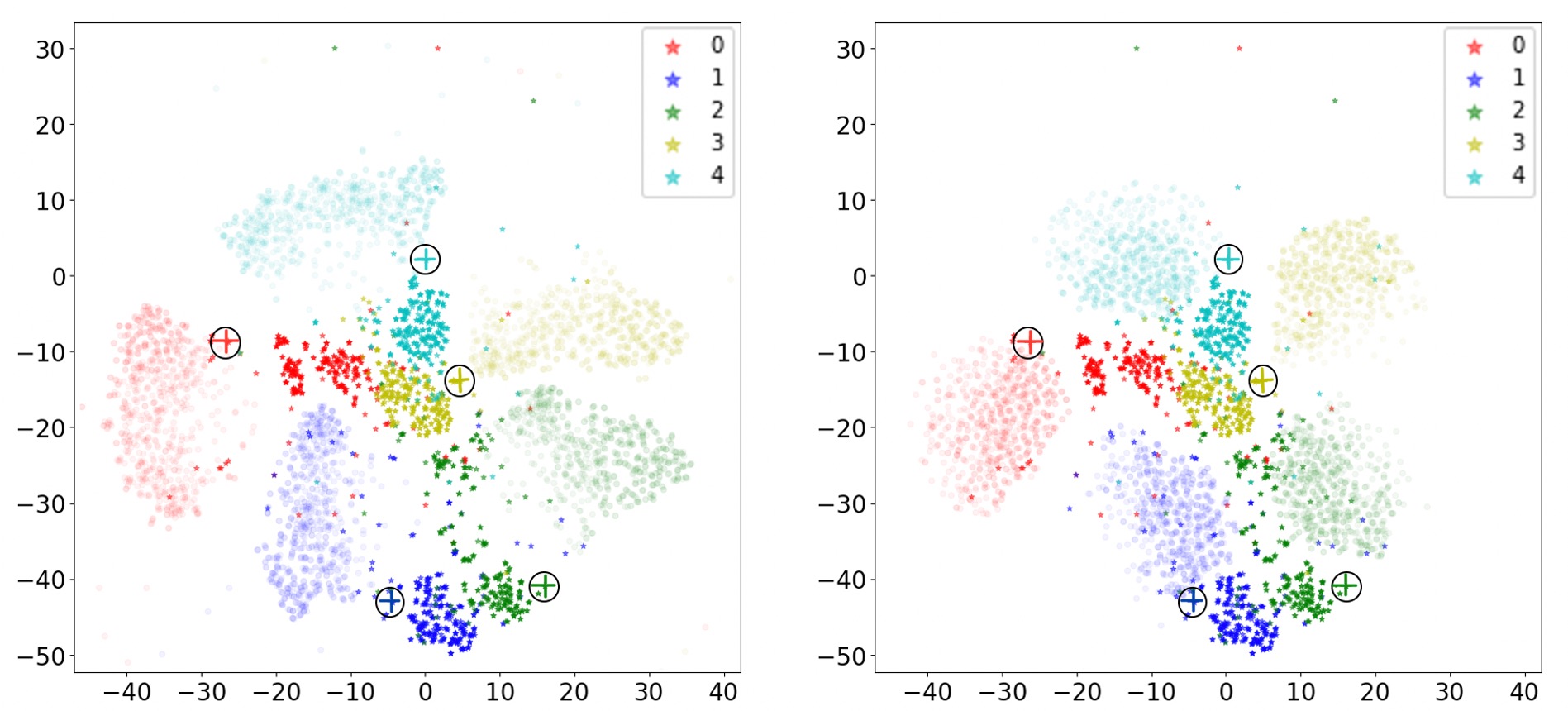}
    \includegraphics[scale=0.18]{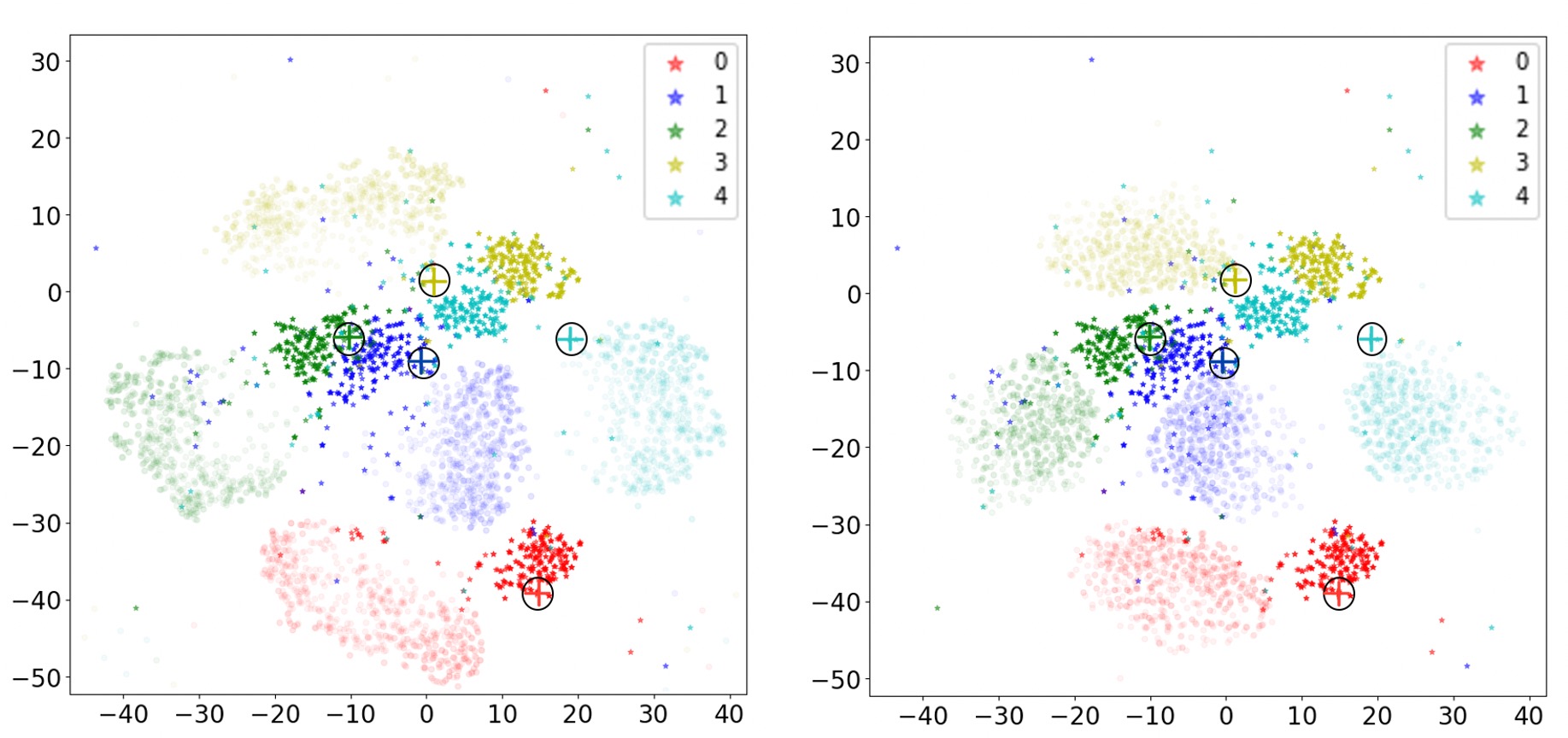}
    \caption{t-SNE visualization for 5 randomly sampled novel classes for DC (left) and our method GDC (right) for miniImagenet, supplementing the visualization in the main text section. The support points are indicated with a '+' sign within black circles, sampled points are semi-transparent indicated with a 'o' sign, and the query points are opaque denoted by '*'. Note that our sampled points ( o ) are on average closer to the ground truth query points ( * ).}
    \label{fig:tsne_3}
\end{figure}

\section{meta-tieredImagenet}
\label{meta-dataset-appendix}


We used the same S2M2 method for training the backbone feature extractor, the feature dimensions were set to 64 with no activation function. Hyperparameter search ranges were, 
\begin{flalign*}
\beta &\in [low=0, high=10, step=0.25],\; &m \in [low=0,  high=3, step=0.25],\\ k &\in [low=2, high=|\sC_b|, step=2] \\
\alpha_1 &\in [0,1000,step=100], &\alpha_2 \in [0, 1000, step=100] \times \alpha_1
\end{flalign*}

In Table \ref{tab:meta-results}, we see that GDC also outperforms DC and baseline (no method) on meta-tieredImagenet even when there is a high dissimilarity between the base and novel/validation classes. The hyperparameters corresponding to these results were,
\begin{flalign*}
    &\text{GDC 5way-1shot}\;\; m=2.25, k=8, \alpha_1=100, \alpha_2=100, \beta=1, n=750 \\
    &\text{GDC 5way-5shot}\;\; m=2, k=10, \alpha_1=100, \alpha_2=100, \beta=1, n=750 \\
    &\text{DC 5way-1shot}\;\; k=2, \alpha=1000, \beta=1, n=750 \\
    &\text{DC 5way-5shot}\;\; k=2, \alpha=900, \beta=1, n=750 
\end{flalign*}

\begin{figure}[]
    \centering
    \includegraphics[scale=0.4]{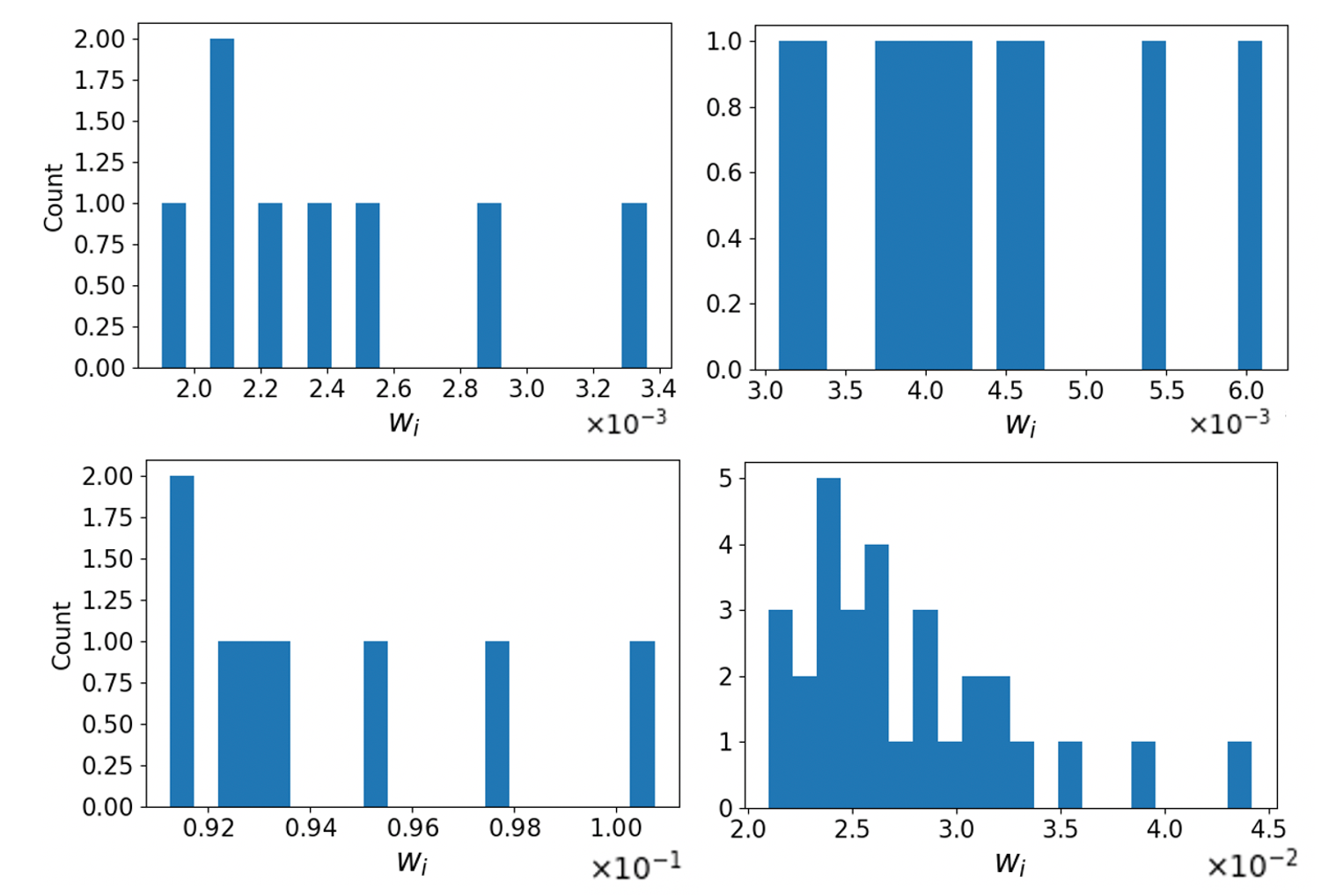}
    \caption{Histogram visualization of weights $w_i$ for meta-tieredImagenet (top) and miniImagenet (bottom) for both 5way-1shot (left) and 5way-5shot (right) tasks. Note how the weights in meta-tieredImagenet are at least an order of magnitude smaller than miniImagenet.}
    \label{fig:wi_histogram}
\end{figure}

An explanation as to why our method outperforms DC on this meta dataset is because of the high $m$ ($m=2.25$ in 5way-1shot and $m=2$ in 5way-5shot) our method discovers. With such high $m$, our $w_i$ is 100 times smaller than the weights assigned by DC method as seen in Figure \ref{fig:wi_histogram}. Hence very small contribution of the $k$ base classes are extrapolated in GDC method while calculating the calibrated $\rvmu', \mSigma'$, compared to DC where a hard contribution of 1/(k+1) exists.

Figure \ref{fig:wi_histogram} and Table \ref{wi_for_k_classes} shows the weights of the $k$ base classes calculated by our method GDC and DC method averaged over 5 random selection of 5way-K shot task (random selection of novel classes and support points)

To put these $w_i$ into perspective, the $w_i$ for best hyperparameters in miniImagenet are 10 times larger as shown in Figure \ref{fig:wi_histogram} and Table \ref{wi_for_k_classes_MI}, ($m=1, k=8$ for 5way-1shot, and $m=3, k=30$ for 5way-5shot) showing that the classes are much more similar in miniImagenet than in the current meta-tieredImagenet.

\begin{table}[]
\caption{$w_i$ for $k$ base classes in meta-tieredImagenet experiment}
\label{wi_for_k_classes}
\begin{center}
\resizebox{\textwidth}{!}{
\begin{tabular}{ll|l}\hline
  Method & 5way-1shot & 5way-5shot \\ \hline
  \multirow{2}{*}{GDC} & [0.0034, 0.0030, 0.0025, 0.0024, 
  & [0.0064, 0.0056, 0.0049, 0.0047, 0.0044, \\
  & 0.0023, 0.0022, 0.0021, 0.0019] & 0.0042, 0.0041, 0.0039, 0.0034, 0.0032] \\ \cline{1-3}
  DC & [0.33, 0.33] & [0.33, 0.33] \\ \hline
\end{tabular}
}
\end{center}
\end{table}

\begin{table}[ht!]
\caption{$w_i$ for $k$ base classes in miniImagenet experiment}
\label{wi_for_k_classes_MI}
\begin{center}
\resizebox{\textwidth}{!}{
\begin{tabular}{ll|l}\hline
  Method & 5way-1shot & 5way-5shot \\ \hline
  \multirow{4}{*}{GDC} & [0.1008, 0.0979, 
  &[0.0442, 0.0393, 0.0355, 0.0332, 0.0323, 0.0318, 0.0307, 0.0303, 0.0298, 0.0291, \\
  & 0.0953, 0.0935, & 0.0287, 0.0281, 0.0273, 0.0267, 0.0264, 0.0261, 0.0258, 0.0255, 0.0253, 0.0250 \\
  & 0.0928, 0.0924,  & 0.0244, 0.0242, 0.0239, 0.0236, 0.0234, 0.0231, 0.0225, 0.0218, 0.0214, 0.0210] \\
  & 0.0916, 0.0912]  & \\ \hline
\end{tabular}
}
\end{center}
\end{table}

\end{document}

%% file: math_commands.tex
\usepackage{amsmath,amsfonts,bm}

\def\eqref#1{equation~\ref{#1}}

\def\1{\bm{1}}

\def\rvx{{\mathbf{x}}}

\def\rvxtilda{\Tilde{\rvx}}
\def\rvmu{{\boldsymbol{\mu}}}

\def\rmI{{\mathbf{I}}}

\def\rmL{{\mathbf{L}}}
\def\rmM{{\mathbf{M}}}

\def\rmX{{\mathbf{X}}}

\def\mSigma{{\bm{\Sigma}}}

\DeclareMathAlphabet{\mathsfit}{\encodingdefault}{\sfdefault}{m}{sl}
\SetMathAlphabet{\mathsfit}{bold}{\encodingdefault}{\sfdefault}{bx}{n}

\def\gN{{\mathcal{N}}}

\def\gT{{\mathcal{T}}}

\def\sC{{\mathbb{C}}}
\def\sD{{\mathbb{D}}}

\def\sQ{{\mathbb{Q}}}
\def\sR{{\mathbb{R}}}
\def\sS{{\mathbb{S}}}

\newcommand{\E}{\mathbb{E}}

